\documentclass[conference]{IEEEtran}
\IEEEoverridecommandlockouts
\usepackage{cite}
\usepackage{amsmath,amssymb,amsfonts}
\usepackage{algorithm}
\usepackage{algorithmic}
\usepackage{graphicx}
\usepackage{textcomp}
\usepackage{xcolor}

\usepackage{tikz}
\usepackage{textcomp}
\usepackage{hyperref}
\usepackage{lipsum}

\newcommand\copyrighttext{%
  \footnotesize \textcopyright 2019 IEEE.  Personal use of this material is permitted.  Permission from IEEE must be obtained for all other uses, in any current or future media, including reprinting/republishing this material for advertising or promotional purposes, creating new collective works, for resale or redistribution to servers or lists, or reuse of any copyrighted component of this work in other works. DOI: \href{<http://tex.stackexchange.com>}{10.1109/IJCNN.2019.8851875}}
\newcommand\copyrightnotice{%
\begin{tikzpicture}[remember picture,overlay]
\node[anchor=south,yshift=10pt] at (current page.south) {\fbox{\parbox{\dimexpr\textwidth-\fboxsep-\fboxrule\relax}{\copyrighttext}}};
\end{tikzpicture}%
}

\def\BibTeX{{\rm B\kern-.05em{\sc i\kern-.025em b}\kern-.08em
    T\kern-.1667em\lower.7ex\hbox{E}\kern-.125emX}}
\begin{document}

\title{Leveraging Recursive Processing for Neural-Symbolic Affect-Target Associations 
\thanks{This project has received funding from the European Union’s Horizon 2020 research and innovation programme under the Marie Skłodowska-Curie grant agreement No 721619 (SOCRATES).}
}

\author{\IEEEauthorblockN{Alexander Sutherland, Sven Magg, Stefan Wermter}
\IEEEauthorblockA{\textit{Knowledge Technology, Department of Informatics,} \\
\textit{University of Hamburg, Germany}\\
Hamburg, Germany \\
$\lbrace$sutherland, magg, wermter$\rbrace$@informatik.uni-hamburg.de}
}

\maketitle
\copyrightnotice
\begin{abstract}
Explaining the outcome of deep learning decisions based on affect is challenging but necessary if we expect social companion robots to interact with users on an emotional level. In this paper, we present a commonsense approach that utilizes an interpretable hybrid neural-symbolic system to associate extracted targets, noun chunks determined to be associated with the expressed emotion, with affective labels from a natural language expression. We leverage a pre-trained neural network that is well adapted to tree and sub-tree processing, the Dependency Tree-LSTM, to learn the affect labels of dynamic targets, determined through symbolic rules, in natural language.  We find that making use of the unique properties of the recursive network provides higher accuracy and interpretability when compared to other unstructured and sequential methods for determining target-affect associations in an aspect-based sentiment analysis task.
\end{abstract}


\section{Introduction}
Within the field of affective computing, the community seeks to learn how to recognise emotions in humans across multiple input modalities and then apply this knowledge to perform emotion-aware decision-making \cite{lim2015recipe}. In spite of a large body of emotion recognition research with promising results \cite{poria2017review}, applications showing how recognised emotions can be applied to interaction scenarios with intelligent agents, such as companion robots, are rare. Emotion recognition is often performed without taking into account the interaction context, interpreting results from the abstract context of the chosen Human-Robot Interaction (HRI)-scenario. From a language perspective, this often entails utterances being associated with emotion labels and robots are to implicitly act without taking into account the granular semantic conversation context.

Understanding how a predicted emotion is reached and how it relates to the specific entities is pivotal for intelligent agents to make decisions surrounding identified emotions, e.g. a subject saying \textit{``You are nice even though you make too much noise.''} should make the agent realise it's too noisy, despite an overall positive sentiment. Given the prominence of neural networks, this also requires the ability to interpret and reason around neural processing and outcome, which is complicated for models that perform high dimensional processing, such as the LSTM \cite{hochreiter1997long}, that process features over a temporal span. Without this ability, intelligent agents lack the understanding to perform even simple emotional reasoning. This can lead to a diminished view of companion robot capability, especially when broaching a topic as intuitive to humans as emotion.

We seek to create interpretable associations that can inform decision-making by understanding how emotions are associated with other semantic elements in natural language. Our approach uses a mixture of commonsense principles, symbolic reasoning, and recursive neural learning to construct associations between affective labels and identified targets in natural language expressions. By adopting principles from similar works \cite{cambria2014senticnet,poria2014sentic} and leveraging features of a specific recursive neural architecture \cite{tai2015improved} we present a system tailored toward HRI-scenarios. The system dynamically identifies multiple \textit{affective targets},  which are noun chunks the system believes the affective expression is aimed towards. It then determines affective language features for the target and provides an affective label for the target.

\begin{figure}[h]
\centerline{\includegraphics[width=0.5\textwidth]{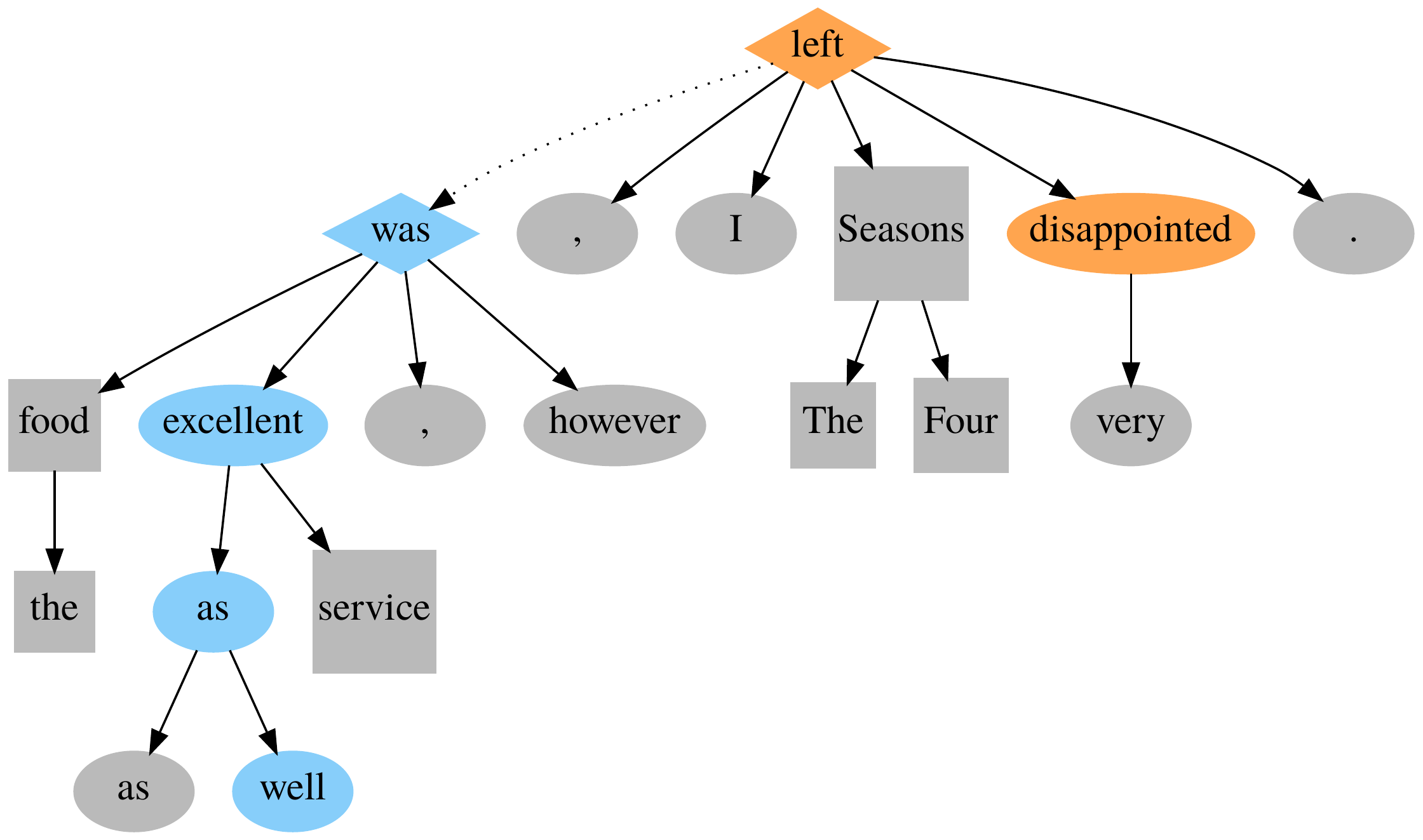}}
\caption{The sentiment parsed dependency tree for the sentence ``\textit{the food was excellent as well as service, however, I left The Four Seasons very disappointed.}'' as provided by a Dependency Tree-LSTM. Coloured nodes indicate that the node acting as the sub-tree of that particular parent node are evaluated as being negative (orange), positive (blue), or neutral (grey). Square nodes are potential targets for the root affective label and the dotted arrow separates two affective feature areas as dictated by root verbs.}
\label{fig}\label{fig:graph1}
\end{figure}

 In this paper, we present a novel method combining symbolic parse-tree rules with recursive neural sub-tree emotion recognition to identify and classify affective labels pertaining to specific targets located within utterances. In our approach, we utilize \textit{sub-trees} of verbs to perform affective labelling of noun chunks in the sub-trees. An example of the type of sub-trees provided by verbs and associated targets can be seen in Figure \ref{fig:graph1}, where the diamond shaped nodes are verbs and the dependency parse can be divided up between the verb sub-trees. In doing so, we are able to leverage the robustness of neural models with the structure and context provided by symbolic rules to construct affect-target pairings without being limited to constraints that are harmful to HRI, such as a predefined number of targets that limit robot speech comprehension or requiring extremely specific datasets.

\section{Related Work}
Contemporary uni-modal and multi-modal emotion recognition has focused primarily on labelling emotion expressions using either continuous or discrete values with late or early fusion\cite{poria2017review}. However, none of these approaches attempt to create associations between sub-parts of language and identified features in a dynamic and proactive manner. Several works have performed Aspect-Based Sentiment Analysis (ABSA), i.e. the correlation of sentiment with some target entity, and often focus on trained approaches toward classifying both aspects and affect \cite{poria2016sentic,he2017unsupervised, hazarika2018modeling, ma2018targeted}. The limitation of these approaches lies with the non-dynamic determination of targets: if a target does not exist within the set of possible or generated target classes then it cannot be considered or identified. This is a problem in the varied and dynamic scenarios companion robots are expected to work within. Our work bypasses the issue of relying too heavily on a predefined set of targets or domain by instead focusing on syntactic rules to select targets.

 Other work also leverages syntactic rules to determine affective targets. Cambria et al. \cite{cambria2015sentic} use a semantic-affective network, SenticNet, to aid in the classification of identified targets. In contrast to their work, our approach labels multiple targets and leverages the compositional processing of recursive neural networks to better classify sub-trees and to retain the parse structure of the network for interpretability. Our system also does not require Sentic patterns \cite{poria2014sentic}, sentiment specific symbolic rules for language interpretation, as we rely solely on the syntactic parse tree to determine targets and their affective areas of interest. As Tree-LSTMs only require root supervision \cite{tai2015improved} and we learn composition rules for different affective labels, our approach is suitable for target-affect extraction across a wide range of labels. This allows us to extend our approach to other affective measurements beyond sentiment, such as valence and arousal \cite{buechel2017emobank}.
 
 

In our previous work \cite{siqueira2018disambiguating}, we had subjects react to audio stimuli and were able to map similar affective reactions to particular stimuli and capture subjective expression styles using an affective association space. We also disambiguated conflicting visual and language expressions by conducting extended dialogue with the subject. While this was an initial first step towards creating a more nuanced understanding of emotion and applying it to a scenario, the learning of user music preferences and ambiguity resolution, we still lack a more nuanced understanding of what it is about stimuli that the users disliked or liked. We now address this by creating affective associations by giving a system the ability to create affective association hypotheses between targets and recognised affective expressions in natural language.

\section{Symbolic Target Identification}
To avoid working with an inflexible set of targets, we assume that affective targets can be derived from syntactic patterns in a dependency parse tree. This is similar to previous work which also defined dependency rules for extracting targets \cite{poria2014sentic}. More specifically we aim to identify noun chunks, which function as targets, that exist as descendants of a verb and, based on the verb, we are able to determine the sub-tree containing relevant affective data for a target. Note, that only the closest parent verb of a target is considered to possess the correct sentiment for a target within its sub-tree. This allows for labelling of multiple targets without the need for target supervision, as we use the verbs as the root for identifying affective sub-trees. By labelling the sub-tree of the verb with an affective label, we are able to classify affective labels for multiple different targets, i.e. noun chunks in the verbs sub-tree, with minimal feature overlap between sub-trees.

To obtain dependency parse trees and noun chunks we use the natural language processing (NLP) library SpaCy\footnote{https://spacy.io/}. Using the dependency parse trees, we identify every verb and obtain the targets associated with that verb's sub-tree which we use as targets. We define the \textit{sub-sentence} of a sub-tree as the in-order sequence of words that can be constructed from that tree. The algorithm for target selection is defined as follows:

\begin{algorithm} 
\caption{Affect Target Identification} 
\label{alg1} 
\begin{algorithmic}[1] 

\STATE {$doc\gets parse(sent)$ }
\FOR{$word$ \textbf{in} $ doc$}
    \IF {$word $ \textbf{is} $ verb$}
        \STATE {$verbs \gets verb$}
    \ENDIF
\ENDFOR
\FOR{$verb$ \textbf{in} $verbs$}
    \STATE {$substring \gets getSubSentence(verb)$} 
    \FOR{$nounChunk$ \textbf{in} $ doc$}
        \IF {$nounChunk$ \textbf{in} $substring$}
            \STATE {$targets \gets nounChunk$}
    \ENDIF
    \ENDFOR
\ENDFOR
\STATE \textbf{return} $targets$
\end{algorithmic}
\end{algorithm}

\subsection{SemEval 2016 Dataset}
There are few evaluation datasets with the desired affect-target relationship we need that also possess an adequate scope of targets. This is due to many datasets focusing on Named Entity Recognition, Topic Recognition, or Aspect Identification to simplify learning and evaluation.  We found that the SemEval 2016 Aspect-Based Sentiment Analysis Data \cite{pontiki2016semeval} is the most appropriate evaluation medium, as it possesses a broad set of aspects with affective sentiment labels.

We extract targets using Algorithm \ref{alg1} and consider the target correct if the labelled aspect exists as a substring in the extracted noun phrase. As an example, the targets ``ball", ``the ball", and ``the red ball" would all be valid extractions for the aspect ``ball". Our symbolic approach is able to identify 1310 aspects out of the possible 1880 (69.7\%) located in the SemEval restaurant data. If we are able to correctly identify an aspect in a sentence, then we attempt to correctly label this aspect using the approached described in Section \ref{TSL}.

\section{Target Sentiment Labelling} \label{TSL}
\subsection{Dependency Tree-LSTMs}
To efficiently determine the affect of a verb sub-tree for a particular target, we utilize a Dependency Tree-LSTM \cite{tai2015improved} trained on the Stanford Sentiment Treebank (SST) \cite{socher2013recursive}. A Tree-LSTM functions, in principle, as the standard LSTM but can take an arbitrary number of inputs at every time-step and processes inputs according to a particular tree-structured format as opposed to in sequence. It also retains the trainable weight matrices $W$ and $U$, as well as a bias vector $b$.

A specific node $j$ in the dependency tree possesses a set of dependent child nodes, denoted $C(j)$, which contains the indexes of children $k$, with each child possessing a hidden state $h_k$. As we are processing dependency trees, the input word vector $x_j$ is the vector corresponding to the word associated with $j$ in the parse tree. The summed hidden states ${\tilde{h}}_{j}$ of the children act as the input to the input gate $i_j$. The Tree-LSTM has multiple forget vectors $f_{jk}$ for each of the $k$ children of a node $j$. Aside from the forget vectors, the Tree-LSTM retains traits of a standard LSTM, such as an output gate activation vector $o_j$ and a memory cell candidate vector $u_j$. The updated memory cell state vector $c_j$ is calculated by summing the elementwise multiplication of $i_j$ and $u_j$ with the sum of the elementwise multiplications of each child cell state $c_k$ with $f_{jk}$. The hidden state $h_j$ is then determined through elementwise multiplication of $o_j$ and $tanh(c_j)$. 

The formalised transition equations for the Dependency Tree-LSTM, as defined in \cite{tai2015improved}, are as follows:

\begin{align}
& {\tilde{h}}_{j} = \sum_{k \in C(j)} h_k, \\
& i_j = \sigma(W^{(i)}x_j + U^{(i)} {\tilde{h}}_{j} + b^{(i)}), \\
& f_{jk} = \sigma(W^{(f)}x_j + U^{(f)} {h}_{k} + b^{(f)}), \\
& o_{j} = \sigma(W^{(o)}x_j + U^{(o)} {\tilde{h}}_{j} + b^{(o)}), \\
& u_{j} = \sigma(W^{(u)}x_j + U^{(u)} {\tilde{h}}_{j} + b^{(u)}), \\
& {c}_{j} = i_j \odot u_j + \sum_{k \in C(j)} f_{jk} \odot c_k, \\
& {h}_{j} = o_j \odot tanh(c_j)
\end{align}

This approach is able to both learn affective language features from a phrase composition perspective and retain an interpretable parse-tree over how a prediction is reached from constituent elements. This compositional and hierarchical processing of input aids in the understanding of how systems reach particular outcomes, as depicted in Figure \ref{fig:graph1}, wherein we see that the two verbs act as the root for two different sentiment interpretations. Note that in practice we would be able to identify additional targets by adding rules that can handle expressions without verbs, i.e. ``Good food.'', however, in this paper, we are only interested in investigating the ability of the system to determine the affective label of a noun chunk existing as a sub-string within the sub-trees we define.

\subsection{Evaluating affective features in sub-trees}
We adapt a PyTorch implementation\footnote{https://github.com/ttpro1995/TreeLSTMSentiment} based on the original implementation \cite{tai2015improved} to perform fine-grained sentiment analysis and we retained the hyper-parameters from the original implementation. We retain the hyper-parameters from the original work for the sake of comparability. As such, the Tree-LSTM had an input of size 300, as dictated by the Common Crawl 840B pretrained embedding, a hidden layer of size 168 and an output layer of size 3 with softmax applied for classification.

As benchmarks against the recursive neural approach we chose to use a Bidirectional LSTM model (BLSTM) approach which has been shown to work well on emotion recognition from text \cite{kim2014convolutional} and a Logistic Regression Linear Classifier, as linear classifiers have been shown to be competitive with neural approaches \cite{schuff2017annotation}. A bag-of-words approach is also less likely to be influenced by sub-sequence data than models normally requiring full sequences. The Logistic Regression approach uses an LMBFGS-solver \cite{byrd1995limited} with a penalty of l2. The BLSTM has a hidden layer size of 64 with a dropout of 0.5 before the final dense layer that applies softmax.

All baseline systems are given the same data as the Tree-LSTM, although they process the data in different non-recursive manners. Our first baseline is a simple Bag-of-Words (BoW) Logistic Regression Classifier, trained on the combination of all sub-trees and labels from the SST, as provided by SST Utils\footnote{https://github.com/JonathanRaiman/pytreebank}. We also provide a benchmark on the BLSTM that processes the SST sub-trees as sequences of word embeddings \cite{pennington2014glove}. For both the LSTM model and the Tree-LSTM we utilize GloVe word embeddings\footnote{https://nlp.stanford.edu/projects/glove/} pretrained on Common Crawl 840B data.


Mapping entire utterances to affective labels has its uses, especially for integration into multi-modal approaches \cite{poria2017review}, this disregards the semantic context of the language. To determine how affective expressions relate to specific entities in language we need a dynamic method of distributing features, i.e. sub-sequences of text, between identified entities. If we do not distribute features we assume that all entities possess the same affective features, which is not necessarily correct.
    
Commonsense principles dictate that the context in which an entity exists will be local to the position of the entity in a sentence. Both LSTMs and 1D-CNNs with a temporal sliding window use this local context to determine what features to extract. These networks have to learn sentence composition and how to interpret sequences of words. While this gives the networks some flexibility, intermediate interpretations of sequences of words can be unclear and difficult to interpret as the system is working in a high dimensional space.
    
Tree-LSTM's recursive and structured processing allows them to learn features from the syntactic structure of language. This allows for the understanding of how sub-tree compositions result in affective labels and it is possible to track the affective reasoning throughout the hierarchical levels of the parse tree. This can be visualised by the coloured nodes in Figure \ref{fig:graph1} showing how the words ``well'' and ``disappointed'' are the beginning of the affective interpretations for the verb roots ``was'' and ``left''. We can also derive that the positive interpretation of the sub-tree ``as well as'' is primarily based on the word ``well''. The final positive interpretation for the entire phrase as depicted by the sentiment value for ``was'' we can also assume is likely reinforced by the word ``excellent''.

\begin{figure*}[h]
\centerline{\includegraphics[width=\textwidth]{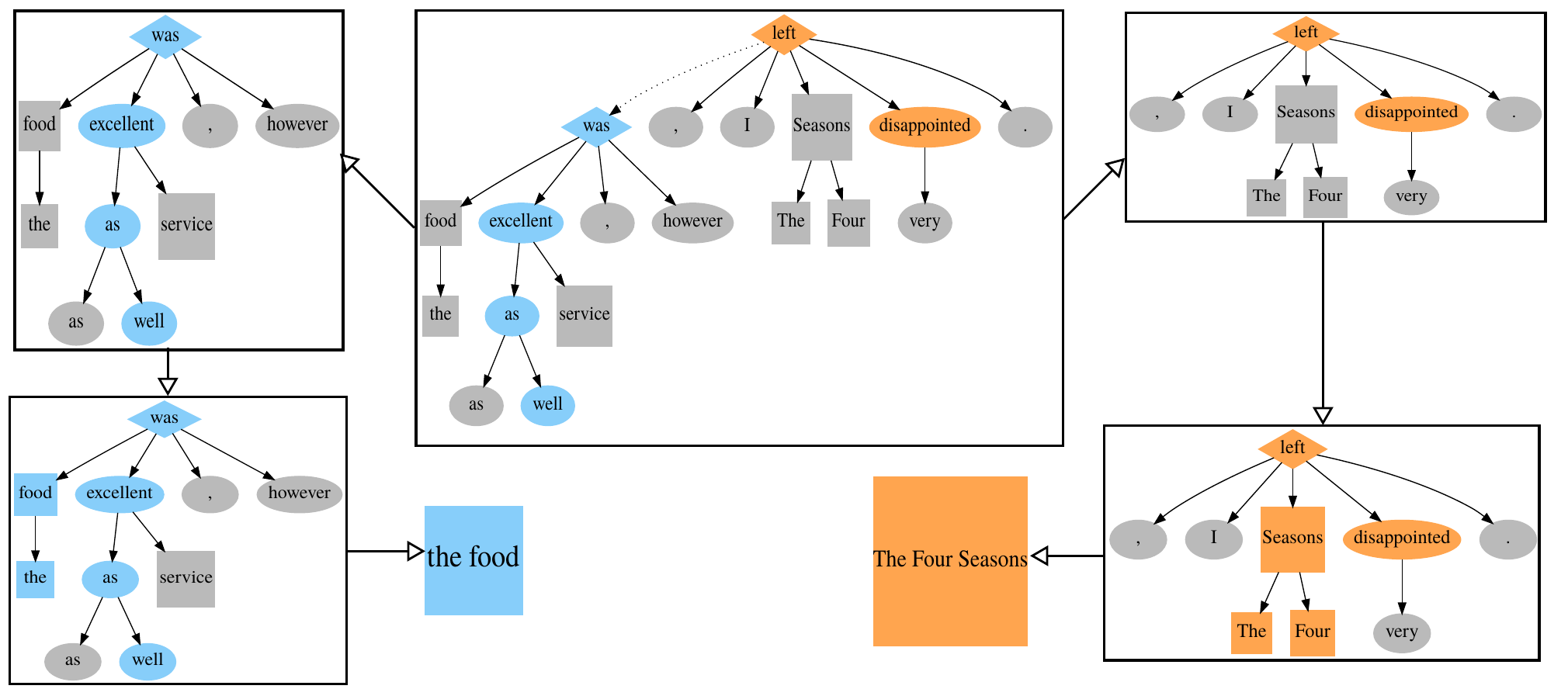}}
\caption{An example of how the affective parse tree provided by the Dependency Tree-LSTM is deconstructed providing a distribution of language features over the set of local targets determined by splitting the tree based on verb phrases, the splitting point here being depicted by a dotted arrow. To the left, we see how positive sentiment trickles down from the root to charge the noun phrase ``\textit{the food}'' and we see how negative sentiment trickles down from the root on the right. Note that although the correct sentiments for ``\textit{the food}'' and ``\textit{The Four Seasons}'' are extracted, the used SpaCy model does not recognize ``\textit{service}'' as a noun chunk and it is therefore not extracted.  }\label{fig:decomp}
\end{figure*}

The selection of dependency sub-trees is based on the principle that affective expressions in English are often described by linking verbs or copula \cite{pavlenko2002emotions}, and we, therefore, divide dependency trees based on their verbs. By selecting the dependency sub-tree associated with a verb we are able to isolate the salient features for the targets within that sub-tree. In Figure \ref{fig:decomp} we see that, by dividing the tree into two sub-trees with the verbs as roots, we are able to capture relevant affective language features to correctly associate emotions with their respective targets ``the food'' and ``The Four Seasons''. 

The process in Figure \ref{fig:decomp} begins with the sentiment parse tree from the Tree-LSTM in the central box, with blue representing positive, grey neutral, and orange negative sentiment. The dotted edge between the two verbs that act as roots is removed to split the tree into two sub-trees with target noun phrases, represented as square nodes in the tree. This results in the top left and top right sub-trees in Figure \ref{fig:decomp}, that have positive and negative roots respectively. The polarity of these roots are then ``trickled down'' into identifiable noun phrases, as seen in the bottom left and right sub-trees in the figure. Finally noun phrases inherit the sentiment from the root of the sub-tree, which is determined to be the sentiment for targets in that sub-tree, resulting in the correct polarities for the targets.


To evaluate how well the Dependency Tree-LSTM works against other models when integrated into our system, we feed the determined sub-trees as Bags-of-Words, sentence order sequences, and sub-trees, to a Logistic Regression model, a Bidirectional LSTM model (BLSTM), and the Dependency Tree-LSTM model respectively. Using these pretrained models we determine whether we can identify the correct sentiment label for the targets we have identified within the sub-tree that was processed. As all models are pretrained on the SST, any biases that would impact their performance on the SemEval will be equal for all models. This still allows for the relative accuracy analysis of sub-tree structured data against other structures. Furthermore, our hybrid method does not focus on the individual statistical performances of the pretrained models on the SST. Instead, we focus on how we can best utilise symbolically determined sub-tree data to identify targets and their affective polarity, with the goal of identifying which model provides the best relative accuracy.


\section{Results}
In this section we show the results for the aforementioned models pre-trained on the SST data and evaluated on the SemEval 2016 ABSA restaurant training data \cite{pontiki2016semeval}.

\begin{table}[h]
\caption{Accuracy on the Stanford Sentiment Treebank for positive, neutral, and negative sentiment for each model. Neural model accuracy is the accuracy for a single fold on validation data.}
\begin{center}
\begin{tabular}{|c|c|c|}
\hline
\textbf{Model} & \textbf{Stanford Sentiment Treebank}  \\
\hline
BoW Logistic Regression & \textbf{0.855}\\ 
\hline
Bidirectional LSTM* & 0.679 \\
\hline
Dependency Tree-LSTM &  0.849 \\
\hline
\end{tabular}
\label{SST}
\end{center}
\end{table}

\begin{table}[h]
\caption{Aspect-based sentiment accuracy performance of the three models over the three labels, pre-trained with the SST dataset, on the SemEval 2016 ABSA restaurant training data \cite{pontiki2016semeval} when disregarding samples without aspects, i.e. out of the correctly identified targets how many are correctly labelled using only sub-tree data.}
\begin{center}
\begin{tabular}{|c|c|}
\hline
\textbf{Model} & \textbf{SemEval 2016 Restaurant Training Data} \\
\hline
BoW Logistic Regression & 0.584\\
\hline
Bidirectional LSTM* & 0.524 \\
\hline
Dependency Tree-LSTM &  \textbf{0.669}\\
\hline
\end{tabular}
\label{SEL}
\end{center}
\end{table}

In Table \ref{SST}, we present a performance comparison between our models on the Stanford Sentiment Treebank dataset \cite{socher2013recursive}. As we can see The Logistic regression model is roughly on par with the Dependency Tree-LSTM, assuming the test set for the neural models is representative. The BLSTM is only trained on the full sentences and root labels of the SST, as it was almost impossible for the network to make use of sub-tree roots due to the high number of redundant samples and class imbalance of the sub-trees \cite{socher2013recursive}.

Results in Table \ref{SEL} indicate that the Dependency Tree-LSTM outperforms the linear classifier when generalising between the SST and SemEval datasets for association determination. The sub-tree data is more descriptive of the target's affective label as opposed to the bag-of-words. This is in spite of the logistic regression model slightly outperforming the Dependency Tree-LSTM on the SST, as seen in Table \ref{SST}. The BLSTM was less efficiently able to learn how to transfer features to the substrings when compared to other models on the SST. 

\subsection{Qualitative association behaviour}

\begin{figure}[h]
\centerline{\includegraphics[width=0.21\textwidth]{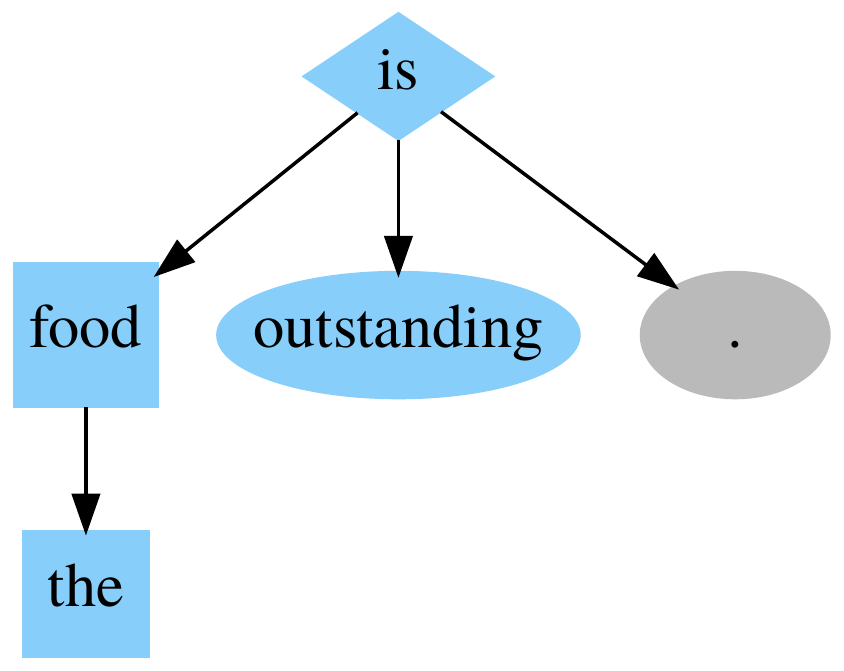}}
\caption{The affective dependency parse tree for  the sentence ``the food is outstanding.'', which indicates a positive sentiment for the target ``the food''. }
\label{fig}\label{fig:ex2}
\end{figure}

One of the core goals of our work is not to just be able to create affect-target associations, but also to have some structured method of understanding why a particular association was created. The reason for this being that social companion agents should reasonably be able to motivate their `train of thought'' regarding a selected action or statement. Take for example Figure \ref{fig:ex2}, where we see that correct sentiment parse for ``the food is outstanding.'' which correctly determines a positive sentiment for ``the food''. Furthermore, if we were to observe the parse before charging ``the food'' with sentiment we would see that the entire tree is uncharged except for ``outstanding'' and ``is''. This allows us to motivate that ``the food'' was viewed positively because ``outstanding'' caused the network to reach that final assertion about the statement.

\begin{figure}[h]
\centerline{\includegraphics[width=0.30\textwidth]{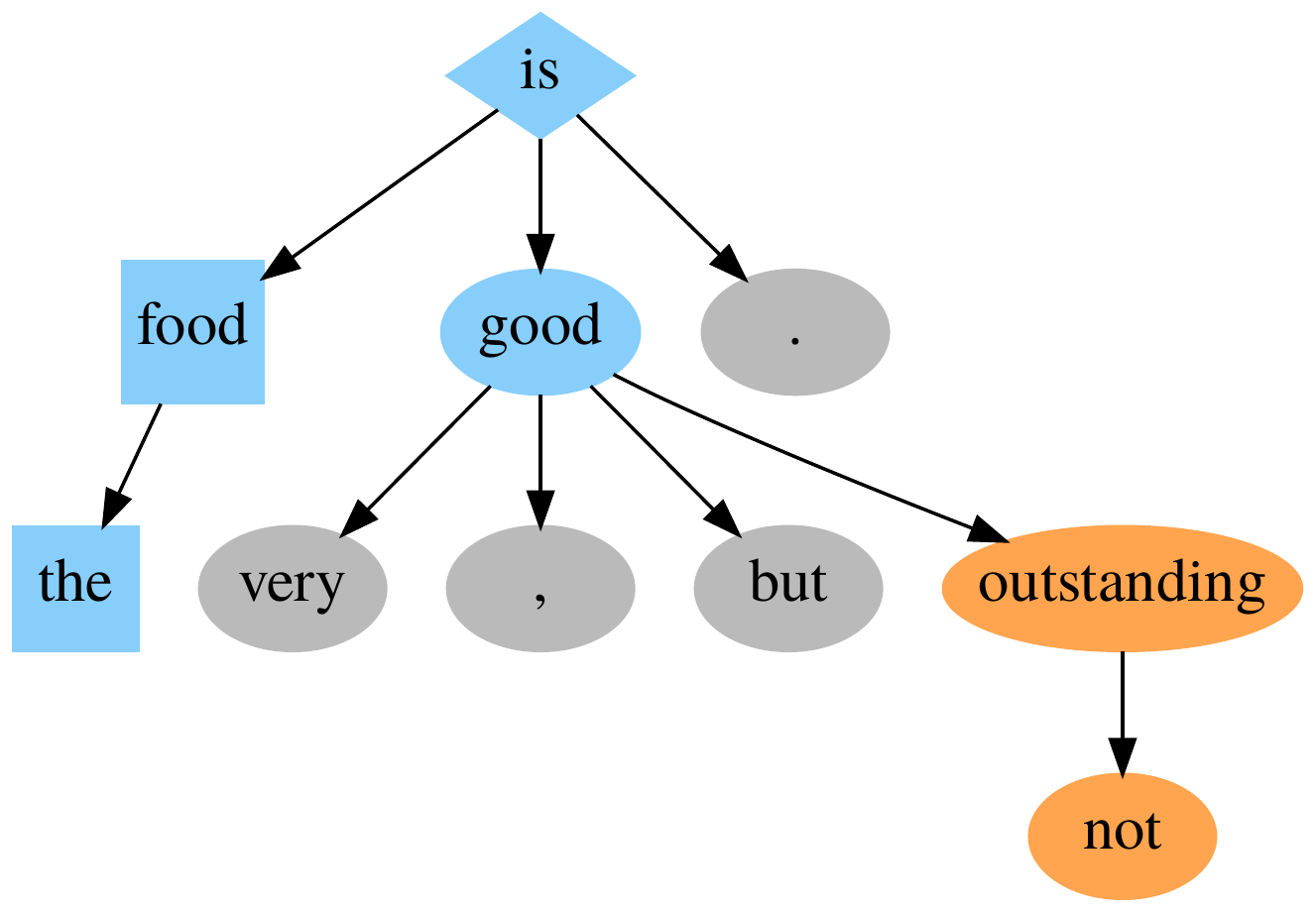}}
\caption{The affective dependency parse tree for  the sentence ``the food is very good, but not outstanding.'', as we can see the negative sentiment from ``not outstanding'' is suppressed by the positive sentiment from the node ``good''.}
\label{fig}\label{fig:ex1}
\end{figure}

In Figure \ref{fig:ex1} we see a contrast to the previous sentence with ``the food is very good, but not outstanding.''. We see the influence of ``not'' on the word ``outstanding'', causing it to be negatively charged. We also see the parent ``good'' overriding the negative sentiment possessed by ``not outstanding'' leading to a final positive inference of ``the food''. While this is labelled incorrect in the data, the data point describes ``food'' as neutral, we would argue that there is a positive sentiment towards ``the food'', as it is described as ``very good''.

\begin{figure}[h]
\centerline{\includegraphics[width=0.21\textwidth]{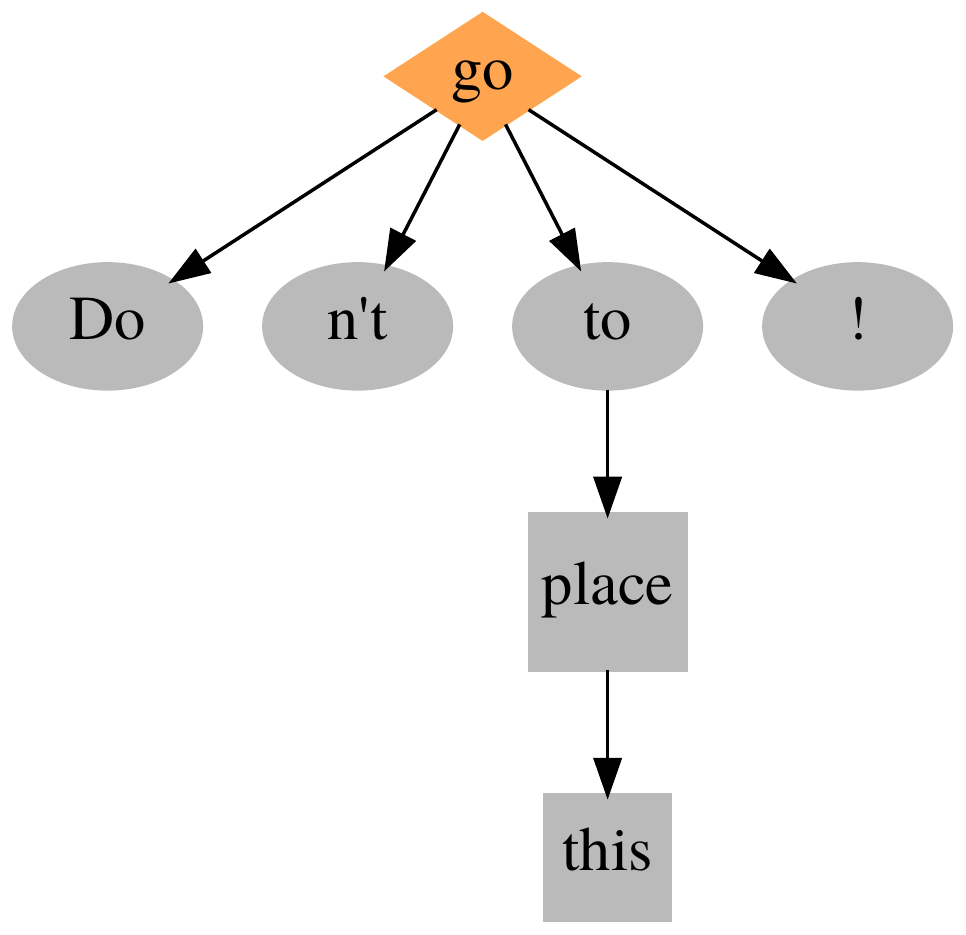}}
\caption{The affective dependency parse tree as provided by our model for the sentence: ``Don't go to this place'' before target association. Our association system will give the correct affective label for ``this place'' but we are unable to reason as to why this is, as all children are deemed to be neutral.}
\label{fig}\label{fig:ex3}
\end{figure}

In Figure \ref{fig:ex3} we see an example of the remaining issue of when none of the children of a particular parent shares the parent's sentiment. We know that the children together lead to a negative sentiment but none of them directly are responsible. This will need to be dealt with in the future if we wish to motivate how certain affective assumptions are made that are outside the scope of our current work.

\section{Discussion}
An argument against this approach toward the creation of emotion-target associations is that the method could be too inclusive of targets. Our approach creates associations from all noun chunks if they are descended from a verb in the dependency parse tree, which also includes determiners. We argue that having a more comprehensive affective understanding of each target in an expression is beneficial, as it provides an intelligent agent with a more nuanced interpretation of the language. Differentiating between affect towards targets as dictated by determiners can provide valuable information to help understand what a user is expressing in context, i.e. talking about the current restaurant owner, ``the restaurant owner is rude'', or describing a new desired restaurant owner ``it would be good to have a more polite restaurant owner''.
    
This is not well suited for aspect-based sentiment analysis from the perspective of opinion mining, due to the high recall but low precision provided by the approach. However, it will provide significantly more data to work with for intelligent agents who are given a wide array of signals and are then able to select which of these is most appropriate to broach in dialogue, or otherwise. Unsupervised approaches \cite{he2017unsupervised} may still provide a dynamic neural solution, as they are able to move away from a fixed set of targets. However, there is still a lack of evidence these approaches toward aspect extraction can reliably extend to multiple domains at once and not suffer the same problems as a symbolic approach. As such, we believe our hybrid neural-symbolic approach is the most suitable for incorporation into current HRI systems, due to its ability to provide affective labels for a dynamic set of targets without compromising the affective labelling.

\section{Conclusion}

In summary, we presented a novel approach of combining the recursive and structured sentiment parse provided by the recursive Tree-LSTM with symbolic rules to perform targeted affect labeling. We identified aspects using our symbolic rule-set for aspect-labelling on the SemEval Restaurant ABSA dataset. Results showed the Dependency Tree-LSTM was best able to utilise the sub-tree features provided by the SST to identify sentiment in sub-trees for individual aspects.  


In future work, we intend to extend upon our previous work \cite{siqueira2018disambiguating} by incorporating the associations we have extracted into HRI-scenarios, allowing for dialogue decisions based on information garnered by our association system to be able to more directly reference emotions in dialogue.

\bibliographystyle{unsrt}
\bibliography{refs}

\end{document}